\documentclass{article}

\usepackage{arxiv}
\usepackage{amsmath}   
\usepackage{amssymb}   
\usepackage[utf8]{inputenc} 
\usepackage[T1]{fontenc}    
\usepackage{hyperref}       
\usepackage{url}            
\usepackage{booktabs}       
\usepackage{amsfonts}       
\usepackage{nicefrac}       
\usepackage{microtype}      
\usepackage{lipsum}
\usepackage{graphicx}
\graphicspath{ {./images/} }

\title{PILOC: A Pheromone Inverse Guidance Mechanism and Local-Communication Framework for Dynamic Target Search of Multi-Agent in Unknown Environments}

\author{
 Hengrui Liu\thanks{School of Control Science and Engineering, Dalian University of Technology, Dalian, China 116024. Email: \texttt{liuhengrui@mail.dlut.edu.cn}} \\
  \And
 Yi Feng\thanks{School of Control Science and Engineering, Dalian University of Technology, Dalian, China 116024. Email: \texttt{fengyi@dlut.edu.cn}. Corresponding author.} \\
  \And
 Qilong Zhang\thanks{School of Control Science and Engineering, Dalian University of Technology, Dalian, China 116024. Email: \texttt{zidonghuazql@163.com}} \\
}

\begin{document}
\maketitle
\begin{abstract}
Multi-Agent Search and Rescue (MASAR) tasks are essential in disaster response, environmental exploration, and military reconnaissance. However, in dynamic and unknown environments, real-time target changes and uncertainties pose significant challenges to traditional MASAR methods. To address these challenges, we propose the PILOC framework, which functions without relying on global prior knowledge. Instead, it leverages agents’ local perception and communication, incorporating a pheromone inverse guidance mechanism to direct cooperative behavior and enable efficient search and localization of dynamic targets.\\
The framework enables collaboration between agents via a local communication mechanism, avoiding reliance on global communication and significantly reducing communication overhead. Unlike traditional methods, the pheromone mechanism is no longer used as an external heuristic but is directly embedded in the observation space of Deep Reinforcement Learning (DRL). This embedding enables indirect collaboration based on environmental perception and helps agents efficiently perform target search tasks in unknown environments. This paper introduces a DRL-based local communication strategy into the control architecture of a multi-agent system and integrates it with the pheromone inverse guidance mechanism. A series of experiments validate its effectiveness in dynamic and unknown environments.\\
Experimental results show that the strategy combining local communication with the pheromone inverse guidance mechanism significantly improves search efficiency and enhances system robustness under dynamic targets and complex environments. It also demonstrates better adaptability and performance compared to traditional methods and other multi-agent algorithms. This study provides new insights for future MASAR tasks and shows strong potential, particularly in communication-constrained and target-dynamic scenarios.
 
\end{abstract}


\section{Introduction}
Multi-agent systems play a vital role in search and rescue, exploration, and other complex tasks. Improving their collaborative efficiency in locating dynamic targets within unknown environments has become a key focus of current research. Traditional methods face several limitations when applied to large-scale, multi-target, and partially observable environments. These include restricted information transmission, high communication overhead, and low collaboration efficiency. Moreover, continuous communication between agents is often impractical in real-world applications. Limited communication range and prolonged dependence on stable links can increase energy consumption and restrict the operational time of search and rescue agents. Agents typically have access only to local information, and the uncertainty of dynamic targets further increases the complexity of SAR tasks. In recent years, Multi-Agent Reinforcement Learning (MARL) has gained significant attention for its strong capability in addressing complex decision-making problems in dynamic, multi-variable environments\cite{bahrpeyma2022review}\cite{agrawal2021multi}. MARL has achieved notable progress in areas such as robotic swarm collaboration\cite{huang2020multi} and unmanned aerial vehicle (UAV) formation control\cite{yuling2023formation}. For example, in robotic swarm control, MARL is widely applied to UAV collaborative navigation and collision avoidance, effectively enhancing task execution in complex environments\cite{gong2023reinforcement}. In multi-agent target search, notable advances have been made using agent observations and joint maps, though most studies still focus on static targets\cite{wakilpoor2020heterogeneous}\cite{zhang2022h2gnn}\cite{tan2022deep}. Therefore, MARL emerges as an effective solution for the complex challenge of efficiently and collaboratively searching for dynamic targets in unknown environments using multiple agents.\\
To address the aforementioned challenges and enhance the efficiency of multi-agent search for dynamic targets in unknown environments, this study draws inspiration from the collaborative foraging behavior of ant colonies through pheromone communication. We propose an innovative framework that integrates a pheromone inverse guidance mechanism into the MASAR system, combined with a distributed local communication strategy. This design significantly improves system performance in terms of search efficiency and robustness.
The main contributions of this paper are summarized as follows:
\begin{itemize}
    \item We propose a dynamic pheromone updating model by introducing the pheromone mechanism from ant colony optimization (ACO) into the MASAR framework. Unlike conventional ACO methods—which increase pheromone concentration to attract agents toward areas of high density—our framework adopts a reverse guidance strategy. Agents release, sense, and update pheromone concentrations along their search paths to avoid redundant exploration and prioritize potential target areas. This innovative mechanism reduces redundant and ineffective exploration, significantly improving target localization efficiency.
    
    \item To address the high energy consumption, limited flexibility, and poor scalability of traditional global communication approaches, we introduce a distributed local communication mechanism. By constraining the communication range and enabling only nearby agents to exchange information, this mechanism minimizes redundant data transfer, lowers overall energy consumption, and greatly enhances the system’s adaptability and scalability.
    
    \item Through comparisons with several state-of-the-art MARL algorithms, our method demonstrates significant improvements in both search efficiency and system robustness. Furthermore, we conduct ablation studies by progressively removing the pheromone inverse guidance mechanism and the local communication module to evaluate their individual contributions. The results confirm the effectiveness and structural rationality of the proposed method.
\end{itemize}
The structure of this paper is organized as follows: Section 2 reviews related work on MARL; Section 3 introduces the problem formulation and dynamic target search and rescue scenarios; Section 4 presents the proposed PILOC framework, including the pheromone inverse guidance mechanism and local communication strategy; Section 5 details the experimental settings, comparative evaluations, ablation studies, and scalability assessments; finally, Section 6 concludes the paper and outlines directions for future work.

\section{Related Work}
\label{sec:headings}
In MASAR tasks, agents are required to autonomously explore unknown terrains, perform environmental mapping, and identify victims to ensure the effective execution of search and rescue operations\cite{niroui2019deep}. MASAR methods are generally categorized into two types: goal-oriented and coverage-oriented approaches\cite{de2023hierarchical}.\\
Goal-oriented methods typically assume that the target location or its probability distribution is known in advance, and the strategy emphasizes rapid and accurate localization of the target area. For instance, Jamshidnejad et al. proposed an adaptive optimal path navigation approach that predicts short-term target locations to focus quickly on high-probability regions\cite{jamshidnejad2018adaptive}. Similarly, Tang et al. developed a distributed algorithm based on approximate policy graphs using the Decentralized Partially Observable Markov Decision Process (Dec-POMDP), enabling efficient search and rescue in dynamic, target-moving scenarios\cite{tang2007complete}.\\
In contrast, coverage-oriented methods aim to comprehensively explore the unknown environment to ensure that all potential targets are discovered. Traditional approaches such as frontier-based algorithms guide agents toward the boundary between explored and unexplored areas. However, these methods may become inefficient in complex maps and are susceptible to redundant exploration\cite{yamauchi1998frontier}. Galceran and Carreras provided a comprehensive review of coverage path planning strategies, highlighting their potential in large-scale, dynamic environments, while also noting limitations in efficiency and adaptability\cite{galceran2013survey}.\\
In addition to path planning and exploration strategies, communication and collaboration are essential components of multi-agent systems. These elements enable agents to work toward a common goal in a coordinated manner. Without effective communication and collaboration, agents cannot share crucial information or synchronize their actions, leading to inefficiencies, increased conflicts, and reduced resource utilization\cite{sun2025multi}. Therefore, communication is not just a means of information exchange; it is, in essence, a manifestation of coordinated behavior among agents\cite{peter2021modern}. Effective coordination ensures the synchronous operation of agents, improves resource utilization, resolves conflicts, and enhances the system's adaptability and robustness. For this reason, optimizing inter-agent communication has become a core issue in system design.\\
In the design of control architectures for multi-agent systems, there are generally two approaches: centralized and decentralized. The centralized approach relies on a central controller to receive state information from all agents, which then computes a trajectory or action for each agent. For example, Wang Y proposed SCRIMP, where agents use an improved, Transformer-based, scalable global communication mechanism for decision-making\cite{wang2023scrimp}. In contrast, the decentralized approach allows each agent to make decisions independently, based on its own and neighboring agents' environmental information. Ayan Dutta proposed an opportunistic communication strategy based on Decentralized Markov Decision Processes (Dec-MDP), aiming to efficiently sample information, maximize environmental data collection, and reduce prediction uncertainty through decentralized collaboration\cite{dutta2021opportunistic}. Both approaches depend on the efficient transfer and communication of state information. As outlined in\cite{farinelli2004multirobot}, communication can be explicit, where messages are routed directly between agents, or implicit, where information is conveyed indirectly through changes in the shared environment\cite{gielis2022critical}.
One of the main challenges of explicit communication in multi-agent systems is finding a balance between scalability and bandwidth. As system size increases, the available bandwidth for each agent decreases, leading to reduced communication efficiency\cite{ghena2019challenge}. This challenge is especially prominent in applications like disaster rescue, where unstable infrastructure introduces issues such as multipath fading, interference from collapsed buildings, and equipment failures, all of which compromise wireless communication reliability\cite{drew2021multi}. Zhang and Lesser addressed this challenge with a coordinated MARL approach for communication-constrained scenarios, introducing a “potential loss function” to dynamically identify key collaboration objects, thus reducing the coordination set size and minimizing the communication burden while maintaining overall performance\cite{zhang2013coordinating}. Consequently, in extreme environments, local communication methods are becoming increasingly important. They reduce the pressure on global communication and ensure efficient agent coordination under harsh conditions\cite{grupen2020low}.\\
With the rise of DRL technology, more research has been applied to MASAR tasks. DRL effectively solves the complex cooperative interaction problem among agents, enhancing adaptability and autonomy in dynamic and unknown environments. Wakilpoor et al. proposed a heterogeneous MARL framework to construct maps and search and rescue strategies in unknown environments, achieving better generalization performance\cite{wakilpoor2020heterogeneous}. Zhang et al. designed a multi-agent exploration method based on a hierarchical graph neural network (H2GNN), which improves collaboration efficiency and search performance in complex, unknown environments\cite{zhang2022h2gnn}. The centralized training and decentralized execution (CTDE) strategy is also widely applied in MASAR tasks. Algorithms such as MAPPO and MADDPG balance collaboration with communication overhead by sharing global information during training and allowing independent decision-making by each agent during execution. Tan et al. developed a decentralized macro-action DRL algorithm that enables agents to explore efficiently and autonomously in complex environments, alleviating the information transfer bottleneck\cite{tan2022deep}.
Although these methods have shown success in static target and partially known environments, they still face greater challenges in the double unknown environment, where both the target and the environment are entirely unknown and dynamically changing. Traditional methods typically rely on predefined targets or maps, which do not adapt effectively to highly dynamic, completely unknown scenarios. Additionally, the energy consumption and communication instability caused by global communication are significant challenges for current multi-agent systems. The innovative framework proposed in this paper, which incorporates a pheromone inverse guidance mechanism and a local communication strategy, effectively addresses these issues, demonstrating excellent search efficiency and scalability in dynamic, unknown environments.

\section{Problem Description}
\subsection{Scene Description}

\begin{figure}
    \centering
    \includegraphics[width=0.25\linewidth]{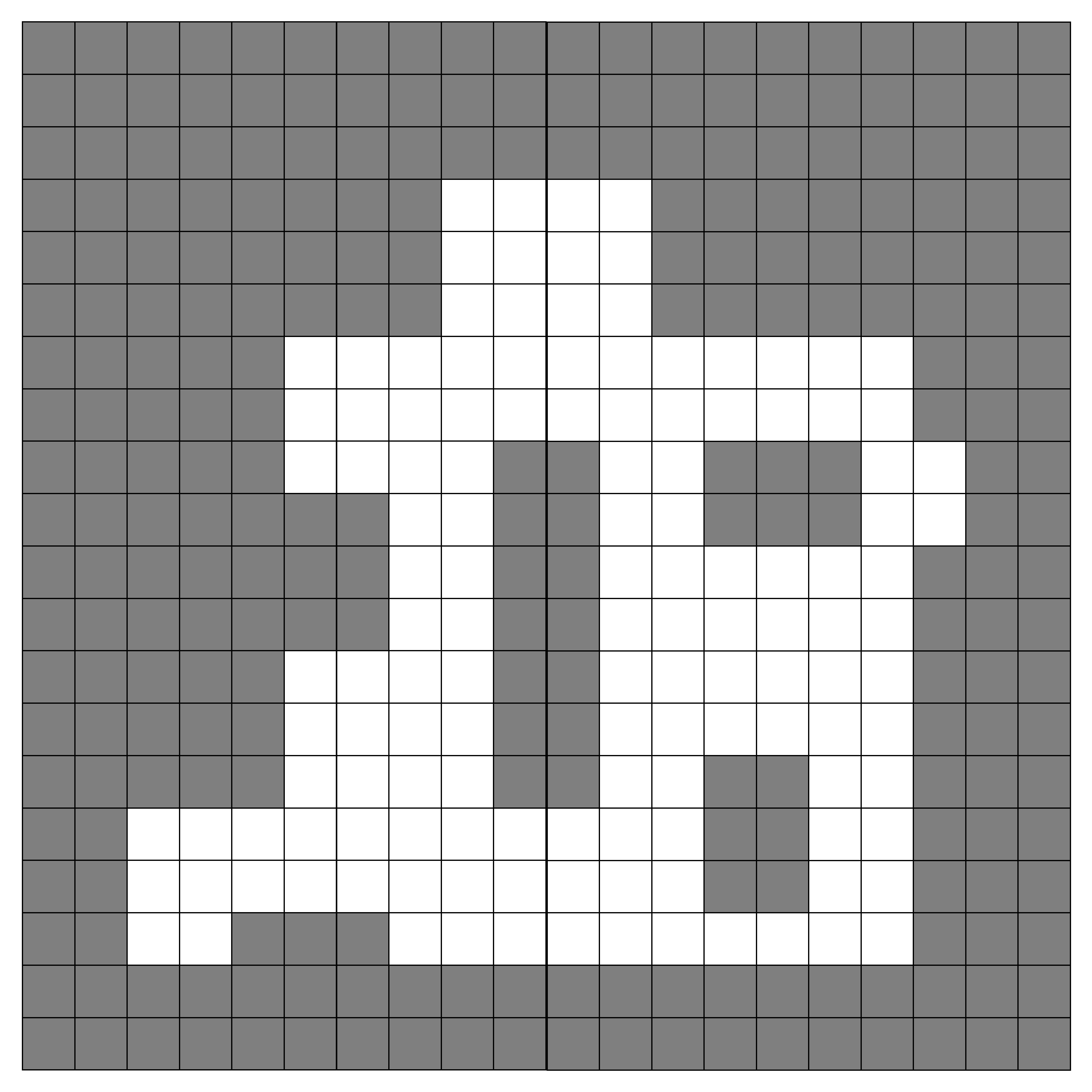}
    \caption{Search and rescue scenario map}
    \label{fig:1}
\end{figure}

We construct a search and rescue scenario on a two-dimensional discretized environment map, as shown in Fig.\ref{fig:1}. The map is composed of gray and white areas, where the gray areas represent impassable regions, such as obstacles, out of bounds areas, and forbidden zones, while the white areas denote passable regions that provide a path selection space for SAR activities. To ensure the feasibility of SAR tasks, we only consider maps in which the passable areas are fully connected. The state of the mission environment is partially observable, meaning that complete information about the entire environment is not directly accessible.
\subsection{Target model}
We consider a dynamic target model with randomly moving search and rescue targets. The SAR scenario consists of multiple targets, denoted as \( G \), which are distributed across the passable area. The initial position of each target \( g_i \in G \) is randomly generated, and at each time step \( t \), target \( g_i \) randomly selects an action \( a_{g_i} \) from the action space \( A_g \). The probability of selecting each action is equal.

\begin{equation}
A_g = \{ \leftarrow, \uparrow, \rightarrow, \downarrow \}
\end{equation}

If target \( g_i \) collides with an obstacle after executing action \( a_{g_i} \), the action is re-selected until an action that does not result in a collision is chosen.

\subsection{Agent Model}

We assume that all agents are homogeneous, and there are \( U \) agents, where \( U \triangleq \{u = 1, 2, \dots, U\} \). Each agent can perceive its surroundings within a perception range \( v \) and communicate with other agents within a communication range \( c \). Agents exchange knowledge gained from their respective explorations, with detailed communication processes discussed in Subsection 4.4. The focus of this work is not on trajectory optimization but on optimizing the search and rescue strategy. To simplify the problem, we discretize the agent's action space into four directions, defined as \( A = \{ \leftarrow, \uparrow, \rightarrow, \downarrow \} \).

Each agent starts randomly from a free cell in a passable area on the map. For each agent, the location of the target to be searched and the locations of other agents (in the absence of communication) are unknown. The goal of the search and rescue task is to locate all the targets within a specified time step.

\section{Method}
\subsection{Problem Modeling and Dec-POMDP Framework}

We describe the MASAR problem in an unknown scenario as a Decentralized Partially Observable Markov Decision Process (Dec-POMDP), which is a generalization of the Markov Decision Process (MDP). A Dec-POMDP consists of an eight-tuple \( \langle N, S, A, O, \Omega, R, T, \gamma \rangle \), where:
\begin{itemize}
    \item \( N \) is the number of agents, which is usually greater than or equal to 2 in multi-agent systems;
    \item \( S \) is the set of all state variables;
    \item \( A = [A_1, A_2, \dots, A_N] \) is the space of joint actions \( a \), where \( A_i \) represents the set of localized actions \( a_i \) that can be taken by agent \( i \);
    \item \( O = [O_1, O_2, \dots, O_N] \) is the set of observations;
    \item \( \Omega(o, s', a) \sim P(o \mid s', a) \) denotes the probability of observing \( o \) in state \( s' \) after taking action \( a \);
    \item \( R(s) \) is the reward function, which returns the immediate reward;
    \item \( T(s, a, s') \sim P(s' \mid s, a) \) denotes the probability of transitioning from state \( s \) to state \( s' \) after taking action \( a \);
    \item Finally, \( \gamma \in [0, 1] \) is the discount factor, which represents the importance attached to future returns.
\end{itemize}

A Dec-POMDP scenario can be described as follows: at each time step, the environment is in some state \( s \in S \). Agent \( i \) can only obtain local observations \( o_i \) and take actions \( a_i \) based on the local observations (or messages delivered by other agents) to change the environment's state from \( s \) to \( s' \) and receive a reward \( r \). The agent's goal is to learn the strategy \( \pi(a_i \mid o_i): O_i \times A_i \to [0, 1] \) in a finite time horizon \( T \) such that the discounted reward is maximized:
\[
G = \sum_{k=0}^{T} \gamma^k r_k
\]
Since the state in a Dec-POMDP is not fully observable, the agent can only receive partial information about the environment. This generalization is more realistic than MDP, making Dec-POMDP a suitable framework for describing SAR tasks.

\subsection{Observational Modeling and Input Representation}

Based on the assumption that all agents are homogeneous, we use agent \( i \) as an example to explain the observation inputs. Each agent's observation consists of three maps: the obstacle map \( M_o \), the exploration map \( M_e \), and the pheromone observation map \( O_{\text{ph}} \), which are described as follows.

\begin{figure}
    \centering
    \includegraphics[width=0.5\linewidth]{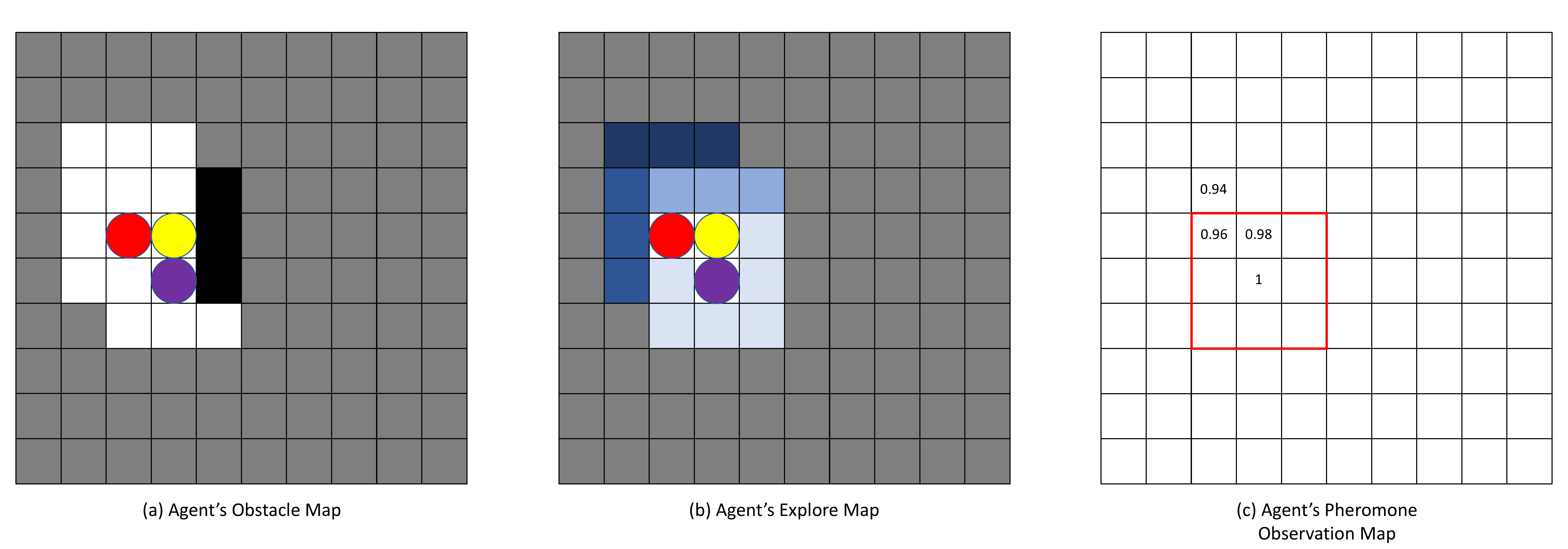}
    \caption{Agent observation inputs: (a) The obstacle map \( M_o \), (b) The exploration map \( M_e \), (c) The pheromone observation map \( O_{\text{ph}} \), where the red box indicates the pheromone observed by the agent.}
    \label{fig2}
\end{figure}
\subsubsection*{Obstacle Map \( M_o \)}

In the obstacle map \( M_o \), four types of data are included, as shown in Fig. \ref{fig2}(a):
\begin{itemize}
    \item \textbf{Unsearched Area}: Gray areas in the obstacle map that may contain obstacles or passable regions.
    \item \textbf{Obstacle Area}: Black areas in the obstacle map, indicating regions that contain obstacles and are impassable.
    \item \textbf{Passable Area}: White areas in the obstacle map, indicating regions that do not contain obstacles and are passable.
    \item \textbf{Historical Position Information}: The obstacle map includes the current position as well as the position information from the previous two time steps.
\end{itemize}

\subsubsection*{Exploration Map \( M_e \)}

Since the target is movable, unlike static targets, it may appear in areas that have already been searched and rescued. The obstacle map does not address this issue, so we introduce the exploration map, as shown in Fig. \ref{fig2}(b).

The exploration map \( M_e \) contains only two types of exploration information, along with historical position data:
\begin{itemize}
    \item \textbf{Unexplored Area}: Gray areas in the exploration map, indicating regions that have not been observed by the agent.
    \item \textbf{Explored Area}: Regions that have been observed by the agent. Note that obstacles are not explicitly shown in the exploration map; it only records which areas have been observed. To indicate the time since the areas were last observed, a time-marking mechanism is introduced. The explored areas are assigned a value between [0, 0.3], which is updated at each time step. A smaller value indicates that the area was observed more recently, and this is represented as a lighter color in Fig. \ref{fig2}(b).
    \item \textbf{Historical Position Information}: The exploration map contains the current position and the position information from the previous two time steps.
\end{itemize}

\subsubsection*{Pheromone Observation Map \( O_{\text{ph}} \)}

To enhance the collaborative efficiency of multiple agents performing search and rescue tasks in unknown environments, we draw inspiration from the communication mechanism of ant colonies and incorporate the Pheromone Inverse Guidance Mechanism into the algorithm as an auxiliary tool.

Each agent considers itself as the center and acquires the pheromone distribution within a fixed range, which forms its pheromone observation map \( O_{\text{ph}} \). Specifically, the size of the pheromone observation window is set to \( l \times l \), where each agent can perceive the current pheromone concentration at all grid points in the square area around itself, with side length \( l \), at each time step, as shown in Fig. \ref{fig2}(c). The pheromone observation map is used as one of the inputs to the neural network, along with the obstacle map and exploration map, for action generation. The specific pheromone mechanism is described in the next section.

\subsection{Pheromone Inverse Guidance Mechanism}

At each time step, the pheromone at the location where the agent is situated is updated, and the pheromone concentration \( P(x,y) \) depends on the time the agent stays at that location and the frequency of visits. To prevent infinite accumulation of pheromone, we set an upper limit for the pheromone concentration, denoted \( P_{\text{max}} \). The pheromone concentration decays gradually at a rate \( \lambda \) over time. The pheromone update rule is as follows:

\begin{equation}
P(x,y) = \begin{cases}
P(x,y) + 1, & \text{Agent exists at this location} \\
P(x,y), & \text{Agent does not exist at this location}
\end{cases}
\end{equation}

If \( P(x,y) > P_{\text{max}} \) after updating, then \( P(x,y) = P_{\text{max}} \).

The pheromone evaporation rule at each time step is as follows:

\begin{equation}
P(x,y) = P(x,y) \times (1 - \lambda)
\end{equation}

where \( \lambda \) is the evaporation rate. In this study, \( P_{\text{max}} \) is set to 10, and \( \lambda \) is set to 0.02. While the standard pheromone mechanism directs the agent toward regions that have been frequently visited, our approach works in the opposite direction by guiding the agent toward less frequently visited regions. The core idea of this mechanism is to prevent ineffective repeated exploration, thereby improving exploration efficiency. Based on this principle, we call it the Pheromone Inverse Guidance Mechanism.
\subsection{Communication Strategies}
Agents can communicate with each other, and when two or more agents are within the communication range \( c \), they will exchange knowledge gained from their explorations to update the obstacle and exploration maps. Assuming there are \( C \) agents communicating, the updated obstacle map \( M_o^{i' } \) of agent \( i \), after applying Equation\ref{eq:4} , is shown on the right side of Fig.\ref{fig3}:
\begin{figure}
    \centering
    \includegraphics[width=0.5\linewidth]{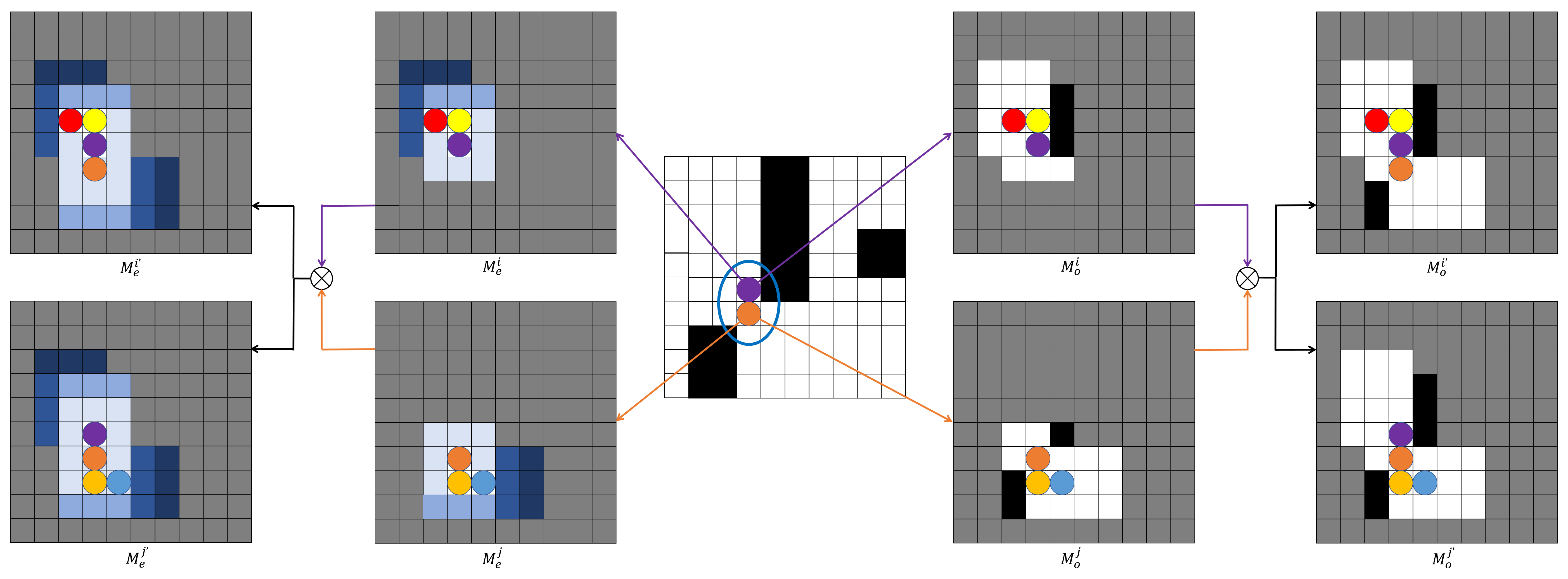}
    \caption{Schematic diagram of agent communication: the left side shows the exploration map update method, and the right side illustrates the obstacle map update method.}
    \label{fig3}
\end{figure}

\begin{equation}
M_o^{(i') } = \bigcup_{j \in C} M_o^j  \quad \forall i \in C \tag{4}
\label{eq:4}
\end{equation}

When updating the exploration map \( M_e^i \), the updated \( M_e^{i' } \) is obtained from Equation\ref{eq:5}, and the update effect is shown on the left side of Fig.\ref{fig3}:

\begin{equation}
M_e^{(i') } = \bigcup_{j \in C} M_e^j  \quad \forall i \in C \tag{5}
\label{eq:5}
\end{equation}

When agent \( i \) and agent \( j \) enter the communication range (indicated by the blue circle), the communication condition is met. Agents \( i \) and \( j \) independently construct their obstacle and exploration maps: \( M_o^i, M_e^i \) and \( M_o^j, M_e^j \), respectively, during the SAR process. Agent \( i \) receives the maps \( M_o^j, M_e^j \) from agent \( j \), and merges them with its own maps \( M_o^i, M_e^i \) to obtain the updated maps \( M_o^{i'}, M_e^{i' } \). Similarly, agent \( j \) synchronously merges the information from agent \( i \) to obtain \( M_o^{j'}, M_e^{j' } \).

When updating the exploration map, due to the time-marking mechanism, it is possible for two agents to have different time markings for the same area. In such cases, the time marking with the smaller value is selected to update the map. This is because the time marking indicates the duration since the area was last explored and is independent of which agent explored it.

\subsection{Neural Network Architecture Design}
\begin{figure}
    \centering
    \includegraphics[width=0.5\linewidth]{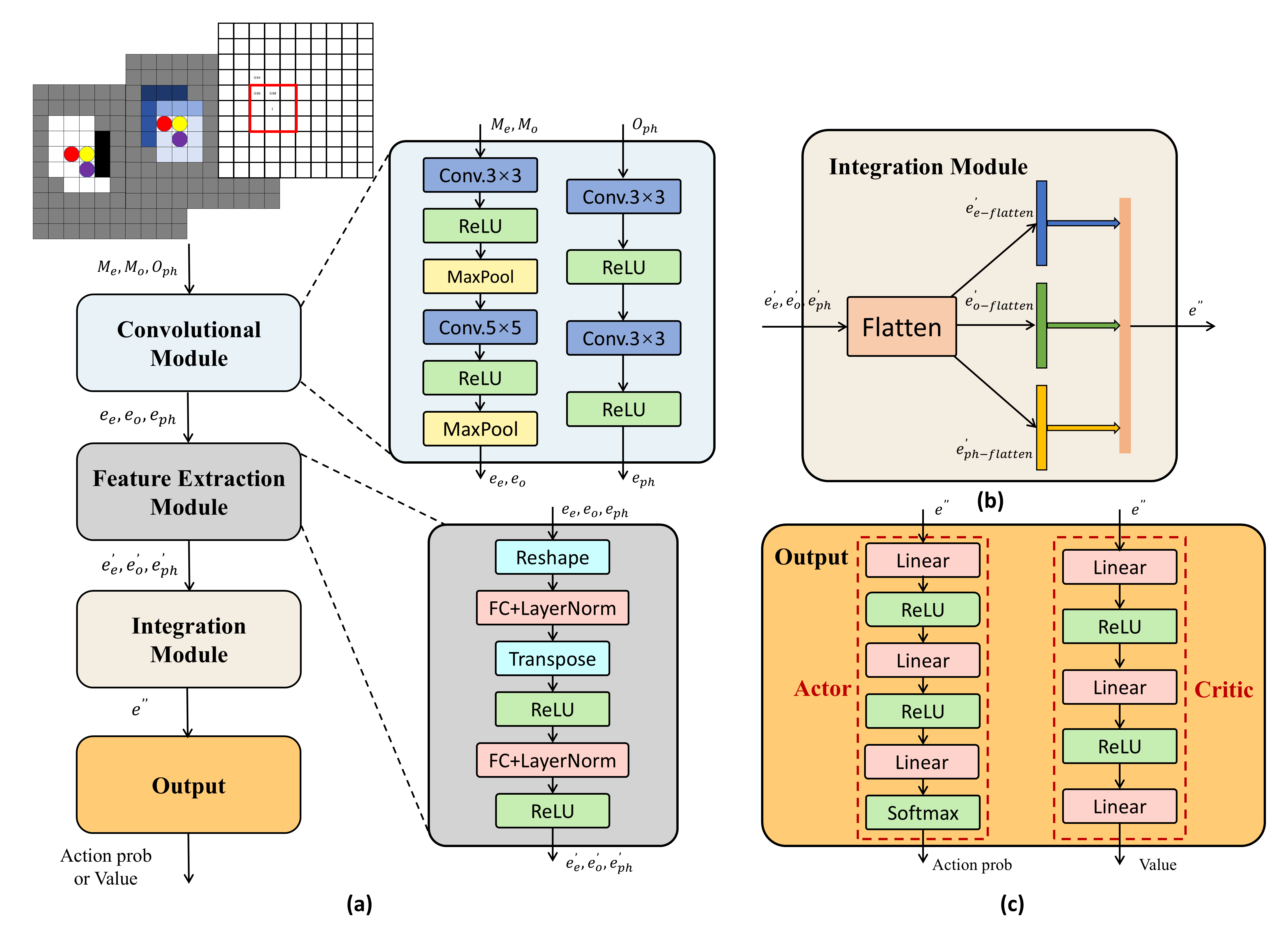}
    \caption{
    (a) Overall architecture of the network, consisting of four main components: the convolutional module, the feature extraction module, the integration module, and the final output layer. 
    (b) Integration module, illustrating the use of channel concatenation instead of weighted fusion. 
    (c) Actor-Critic output structure, showing the separate roles of the actor and critic components in action selection and value estimation, respectively.
    }

    \label{Fig4}
\end{figure}

To process the input information from the three observation maps—namely, the obstacle map \( M_o \), the exploration map \( M_e \), and the pheromone observation map \( O_{\text{ph}} \)—we construct a neural network framework based on the Actor-Critic architecture. 
As illustrated in Fig.\ref{Fig4}, the network consists of four components: a convolutional module, a feature extraction module, an integration module, and an output layer.

For each input map, spatial features are progressively extracted through a series of convolution and pooling operations in the convolutional module. The resulting feature maps are reshaped and passed through a fully connected (FC) layer, which integrates and transforms the channel-wise features while capturing the inter-channel dependencies. The reshaped features are then transposed to adjust their dimensions and are further processed by another FC layer along the new axis to enrich the representation.\\
The features extracted from all three maps are then fused via the integration module to form a unified and comprehensive feature representation. When the network functions as an actor, the output layer produces a probability distribution over the possible actions, from which an action is selected via probabilistic sampling. Conversely, when the network serves as a critic, the output layer computes the state value through a fully connected (FC) layer, providing a value estimate for the current state to guide decision-making.

\subsection{Reward Shaping}
To improve the agent’s ability to search for and rescue dynamic targets in unknown environments, we design the following reward function:

\begin{equation}
r = r_e + r_{re} - r_{co} + r_{ph} \tag{6}
\end{equation}

where:
 \( r_e \) is the exploration reward. To encourage agents to explore unknown areas, we define \( n_e \) as the number of new passable grid cells discovered by the agent when transitioning from state \( s \) to \( s' \). The exploration reward is calculated as:

\begin{equation}
r_e = 0.5 \times n_e \tag{7}
\end{equation}

 \( r_{re} \) is the re-exploration reward. Since targets are dynamic and may reappear in areas already explored, we design this reward to incentivize agents to revisit previously explored regions. It is defined as:

\begin{equation}
r_{re} = \begin{cases} 
0.01 \times \sum_{i=0}^{n} \left(1 + \frac{v_{m}}{0.3}\right) - 1, & \text{when agent doesn’t find the target} \\
0.1, & \text{when agent finds the target} 
\end{cases} \tag{8}
\end{equation}

where \( n \) is the number of passable grids within the agent's current field of view, and \( v _{m}\) is the time-marking value of each grid. Regions with higher time markings (indicating a longer duration since the last observation) are considered more valuable for re-exploration.

 \( r_{co} \) represents the collision penalty. Each time an agent collides with an obstacle, a fixed penalty of \( -3 \) is applied to discourage such behavior.

 \( r_{ph} \) is the pheromone reward, which promotes the agent’s use of the Pheromone Inverse Guidance Mechanism. Agents are encouraged to move toward areas with lower pheromone concentrations, thus improving search coverage. The pheromone reward is defined as:

\begin{equation}
r_{ph} = \alpha \frac{I_{\text{ph}}' - I_{\text{ph}}}{I_{\text{ph}}'} + \beta (I_{\text{ph}}' - I_{\text{ph}}) \tag{9}
\end{equation}

where \( I_{\text{ph}}' \) and \( I_{\text{ph}} \) are the pheromone concentrations in the perceptual range at the previous and current time steps, respectively. This formulation considers both relative and absolute changes to improve sensitivity to local variations while maintaining reward stability.In this study, we set $\alpha=0.1, \beta=0.1$.

\subsection{Learning Method}
The performance of single-agent reinforcement learning (RL) algorithms in multi-agent systems is often suboptimal, as an individual agent’s behavior is influenced not only by the environment but also by the actions of other agents. This interdependence introduces non-stationarity from the perspective of each agent, making it difficult to converge using traditional single-agent RL approaches. Therefore, it is necessary to adopt multi-agent reinforcement learning (MARL) techniques to train agents effectively within a Decentralized Partially Observable Markov Decision Process (Dec-POMDP) framework.
\begin{figure}
    \centering
    \includegraphics[width=0.5\linewidth]{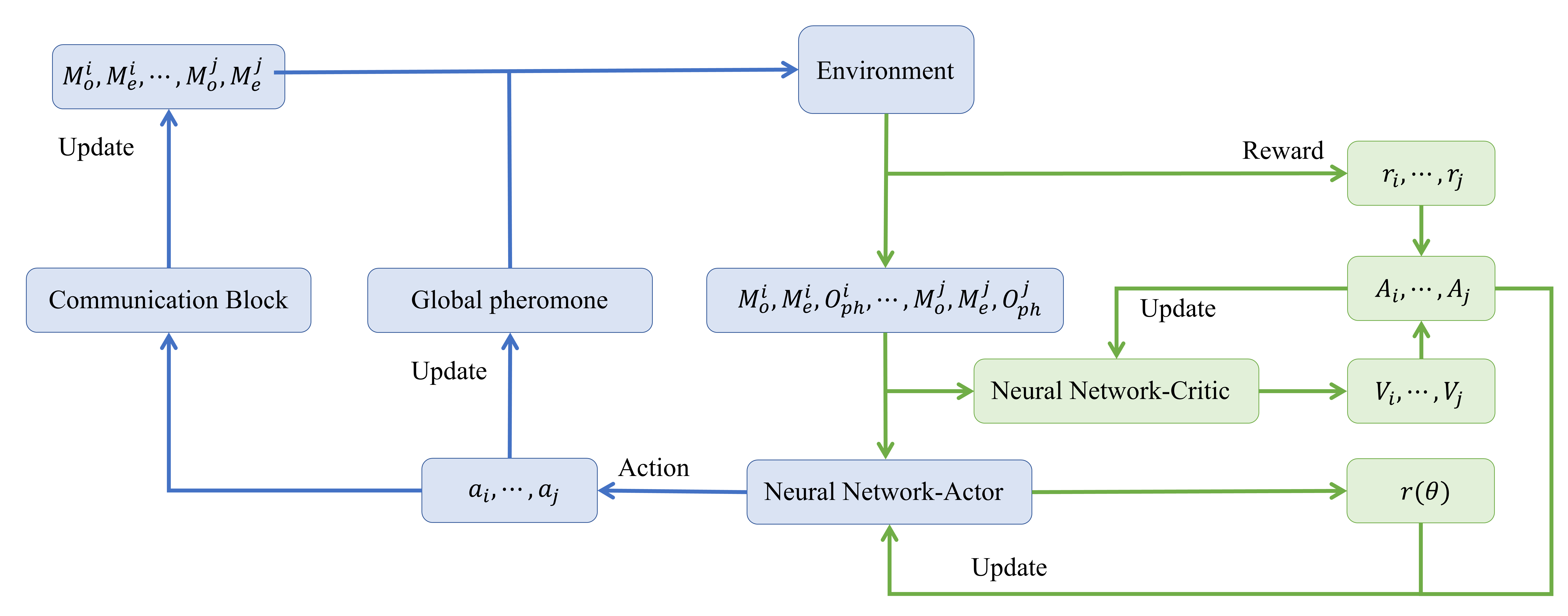}
    \caption{The PILOC framework proposed in this paper, which combines local communication and pheromone mechanism, illustrates the update path of the neural network and the interaction process with the environment. Herein, the blue part represents the interaction between the agent and the environment, and the green part represents the update of the neural network.}
    \label{fig:5}
\end{figure}
In this study, we employ Multi-Agent Proximal Policy Optimization (MAPPO), an advanced variant of the Proximal Policy Optimization (PPO) algorithm, as our learning method,as shown in Fig.\ref{fig:5}. MAPPO leverages the CTDE paradigm, which enables agents to share global information during training for improved learning efficiency, while ensuring independent decision-making during execution. This architecture is particularly suitable for scenarios with limited communication or computational resources.

PPO itself is a policy-gradient algorithm based on the Actor-Critic framework and is widely recognized for its stability and ease of implementation. It was developed as an improvement over Trust Region Policy Optimization (TRPO), addressing the limitations related to step-size sensitivity and optimization complexity in TRPO. PPO introduces a clipped surrogate objective that restricts excessive updates to the policy, thereby maintaining training stability. The learning objective of reinforcement learning is to optimize a policy \( \pi^* \) that maximizes the expected cumulative discounted return \( R_t \), formulated as:

\begin{equation}
R_t = r_{t+1} + \gamma r_{t+2} + \gamma^2 r_{t+3} + \cdots = \sum_{k=0}^{\infty} \gamma^k r_{t+k+1} \tag{10}
\end{equation}

Where \( \gamma \in [0, 1] \) is the discount factor. The policy \( \pi \) can be evaluated using the state-value function \( V^{\pi}(s) \) and the action-value function \( Q^{\pi}(s, a) \). The state-value function represents the expected return when following policy \( \pi \) starting from state \( s \):

\begin{equation}
V^{\pi}(s_t) = \mathbb{E}_{\pi} [R_t | s_t] = \mathbb{E}_{\pi} \left[ \sum_{k=0}^{\infty} \gamma^k r_{k+1} | s_t \right] \tag{11}
\end{equation}

The action-value function \( Q^{\pi}(s, a) \) represents the expected return when taking action \( a \) in state \( s \) under policy \( \pi \):

\begin{equation}
Q^{\pi}(s_t, a_t) = \mathbb{E}_{\pi} [R_t | s_t, a_t] = \mathbb{E}_{\pi} \left[ \sum_{k=0}^{\infty} \gamma^k r_{k+1} | s_t, a_t \right] \tag{12}
\end{equation}

To improve the training stability and efficiency, the advantage function \( A^{\pi}(s, a) \) is introduced:

\begin{equation}
A^{\pi}(s_t, a_t) = Q^{\pi}(s_t, a_t) - V^{\pi}(s_t) \tag{13}
\end{equation}

This function measures how much better taking action \( a \) in state \( s \) is compared to the average expected return under policy \( \pi \). The estimated advantage at timestep \( t \), denoted \( \hat{A}_t \), is given by:

\begin{equation}
\hat{A}_t = -V(s_t) + r_t + \gamma r_{t+1} + \cdots + \gamma^{(T-t+1)} r_{T-1} + \gamma^{(T-t)} V(s_T) \tag{14}
\end{equation}

where \( T \) is the trajectory length. Policy gradient methods optimize policies directly, and the clipped surrogate objective function in PPO is defined as:

\begin{equation}
L^{\text{CLIP}}(\theta) = \mathbb{E} \left[ \min \left( r_t(\theta) \hat{A}_t, \text{clip}(r_t(\theta), 1-\epsilon, 1+\epsilon) \hat{A}_t \right) \right] \tag{15}
\end{equation}

where \( r_t(\theta) = \frac{\pi_{\theta}(a_t | s_t)}{\pi_{\theta_{\text{old}}}(a_t | s_t)} \) is the probability ratio of the new and old policies. The clip function limits the policy update to a constrained region, thereby preventing large and potentially unstable policy shifts.

In this study, MAPPO is employed to train two separate networks: the policy network \( \pi_{\theta} \) with parameters \( \theta \), and the value network \( V_{\phi} \) with parameters \( \phi \). All homogeneous agents share these networks and adopt a common policy during training and inference, which ensures both consistency and convergence. The policy network maps agent observations to actions, while the value network estimates the expected return given the state.

In the multi-agent setting, the clipped objective function for the policy network becomes:

\begin{equation}
L^{\text{CLIP}}(\theta) = \frac{1}{BN} \sum_{i=1}^B \sum_{k=1}^N \min \left( r_{\theta, i}^{(k)} \hat{A}_i^{(k)}, \text{clip}(r_{\theta, i}^{(k)}, 1-\epsilon, 1+\epsilon) \hat{A}_i^{(k)} \right) \tag{16}
\end{equation}

where \( B \) is the batch size, \( N \) is the number of agents, \( r_{\theta, i}^{(k)} = \frac{\pi_{\theta}(a_i^{(k)} | s_i^k)}{\pi_{\theta_{\text{old}}}(a_i^{(k)} | s_i^k)} \), and \( \hat{A}_i^{(k)} \) is the estimated advantage.

The corresponding loss function for the value network is:

\begin{equation}
L^{\text{CLIP}}(\phi) = \frac{1}{BN} \sum_{i=1}^B \sum_{k=1}^N \max \left[ \left( V_{\phi}(s_i^{(k)}) - \hat{R}_i \right)^2, \left( \text{clip}(V_{\phi}(s_i^{(k)}), V_{\text{old}}(s_i^{(k)}) - \epsilon, V_{\text{old}}(s_i^{(k)}) + \epsilon) - \hat{R}_i \right)^2 \right] \tag{17}
\end{equation}

\subsection{Hybrid Decision Mechanism}
The hybrid decision mechanism designed in this paper is enabled in the testing phase as a condition-triggered remediation strategy that combines neural networks with traditional search and rescue algorithms. During environment interaction, when agent \( i \) revisits its position more than 3 times within the last 10 time steps, the system determines that the agent is trapped in a localized area due to ineffective exploration. At this point, it switches from neural network-based decision making to a rule-based SAR algorithm for action generation.

Specifically, the algorithm identifies the nearest unexplored grid cell \( p_i \) from the current location of agent \( i \). If all grid cells have already been explored, the system selects the nearest passable cell with the largest time-marking value as the new target, encouraging the agent to explore areas that have not been visited for a long time. The \( A^* \) algorithm is then employed to compute the optimal path from the current location to \( p_i \), and the agent follows this path to continue the exploration.

This hybrid decision approach is inspired by prior works such as HMA-SAR \cite{cao2024hma}, where similar rule-based recovery strategies were adopted to avoid agents becoming trapped or performing suboptimal behaviors in completely unknown environments. Drawing on this idea, the proposed mechanism enhances robustness and exploration completeness, especially under dynamic target conditions.

\section{Experiments}

\subsection{Experimental Setting}
The PILOC framework was trained and tested on the publicly available dataset \cite{chen2019self}, which consists of 5663 grid world maps for training and 5218 for testing. Each grid map is sized \(60 \times 60\). The perception radius of each agent is set to 5.

To improve learning stability and efficiency, a curriculum learning strategy was adopted during training. Specifically, the initial maximum number of time steps per episode \( N_s \) was set to 10, and a threshold \( M \) was used to determine the progress of learning. If the maximum return \( R_{\text{max}} \) did not improve for \( M \) consecutive training episodes, \( N_s \) was increased by 10. This process continued until \( N_s \) reached 260, at which point the training was terminated. This curriculum learning mechanism allowed the agent to learn from simple tasks to more complex ones progressively, which helped avoid instability in the early training stage and promoted better generalization in later stages.

The entire training process was conducted on a desktop equipped with an Intel Core i7-10700 @ 2.90GHz (8-core) CPU and an NVIDIA GeForce RTX 3070 GPU.

For evaluation, 250 scenarios were randomly selected from the test set. In each scenario, six dynamic targets and two agents were randomly initialized. The agents and targets move simultaneously at each time step. Targets do not collide with obstacles or cross boundaries. Agents share identical policy network parameters during both training and testing. The maximum number of steps per test episode is 250; if not all targets are found within this limit, the episode is considered a failure.

\subsection{Comparison Experiments}
To evaluate the effectiveness of the proposed PILOC model, we compare it with several baseline methods, including MASAC \cite{lowe2017multi}, IPPO \cite{de2020independent}, QMIX \cite{rashid2020monotonic}, and the classical rule-based Frontier method \cite{yamauchi1998frontier}. We define the following evaluation metrics:
\begin{itemize}
    \item \textbf{Success Rate (SR)}: The percentage of test episodes in which all targets are successfully located, indicating the overall success stability and robustness of the algorithm.
    \item \textbf{Average Steps (AS)}: The average number of time steps required to complete the search task across 250 test maps, reflecting the efficiency of exploration and execution.
    \item \textbf{Step Variance (SV)}: The variance of steps taken in successful rounds, representing the consistency of performance across different environments.
    \item \textbf{Average Number of Targets Obtained (ANTO)}: The average number of dynamic targets found in each episode, including failed episodes, which reflects the agent’s partial task completion ability.
\end{itemize}

The quantitative results are shown in the table below:
\begin{table}[h!]
\caption{Performance Comparison of Different Algorithms on the Dynamic Target Search and Rescue Task.}
\centering
\begin{tabular}{lcccc}
\toprule
\textbf{Algorithm} & \textbf{SR} & \textbf{AS} & \textbf{SV} & \textbf{ANTO} \\
\midrule
PILOC & 95.6\% & 129.08 & 2897.15 & 5.716 \\
IPPO  & 87.2\% & 158.00 & 3190.97 & 5.148 \\
MASAC & 74.0\% & 174.80 & 2995.33 & 4.344 \\
QMIX  & 62.8\% & 189.64 & 3232.90 & 5.40  \\
Frontier & 18.4\% & 225.14 & 4948.43 & 3.82  \\
\bottomrule
\end{tabular}
\label{tab:performance_comparison}
\end{table}
The experimental results demonstrate that PILOC significantly outperforms all baseline methods. It achieves the highest success rate of 95.6\%, considerably surpassing IPPO (87.2\%), MASAC (74.0\%), QMIX (58.4\%), and Frontier (18.4\%). Additionally, PILOC has the lowest average number of steps and highest average number of targets obtained, indicating superior efficiency in locating dynamic targets. The lowest step variance also reflects its robustness and generalization across diverse SAR environments.

\subsection{Ablation Experiments}
To evaluate the individual contributions of the Pheromone Inverse Guidance Mechanism and the Local Communication Strategy, we conducted a series of ablation experiments under consistent curriculum training and test settings. Three experiment groups were designed as follows:
\begin{itemize}
    \item \textbf{PILOC-com-ph}: the baseline setting, where both the Pheromone Inverse Guidance Mechanism and local communication were removed;
    \item \textbf{PILOC-ph}: local communication was enabled, but the Pheromone Inverse Guidance Mechanism was disabled;
    \item \textbf{PILOC-com}: the Pheromone Inverse Guidance Mechanism was enabled, but local communication was disabled.
\end{itemize}

\begin{figure}
    \centering
    \includegraphics[width=0.4\linewidth]{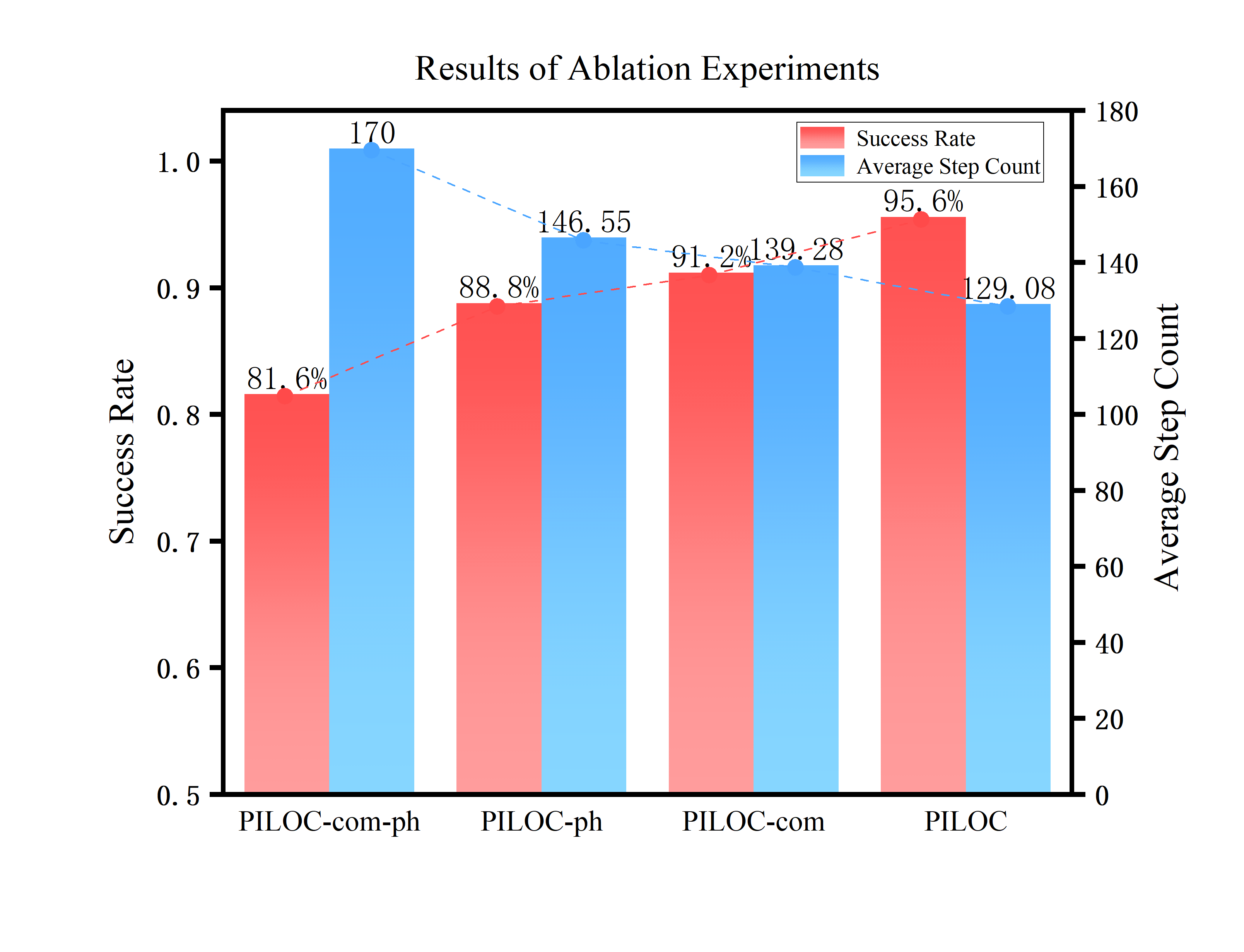}
    \caption{Results of the ablation experiment: shows the comparison results of task success rates (red bars) and average steps (blue bars) under different combinations of functional modules (PILOC-com-ph, PILOC-ph, PILOC-com, and the complete model PILOC).}
    \label{fig:6}
\end{figure}

As illustrated in Fig.\ref{fig:6} the baseline model PILOC-com-ph achieved the lowest success rate (81.6\%) and the highest average number of steps (170), indicating that without any collaborative mechanisms, agents are prone to inefficient behaviors such as redundant exploration and poor coverage. When the Local Communication Strategy is introduced (PILOC-ph), agents can share local map information to reduce duplicate exploration, resulting in a notable improvement in success rate (88.8\%) and a reduction in average steps (146.55). Similarly, enabling the Pheromone Inverse Guidance Mechanism (PILOC-com) without communication still allows agents to perform indirect coordination by observing pheromone concentration. This configuration yields a success rate of 91.2\% and an average of 139.28 steps, showing that the pheromone mechanism effectively guides agents toward unexplored regions and mitigates exploration redundancy.The complete model PILOC, which combines both mechanisms, demonstrates the best performance across all metrics, verifying the complementary and synergistic nature of the pheromone-based and communication-based strategies.

\subsection{Study of Scalability}
In addition, we conducted experiments to investigate the scalability of the PILOC model. Two evaluation metrics were selected: Success Rate (SR) and Average Steps (AS). The experimental setup remains consistent with Section 5.1, with the only variable being the number of agents, which increased incrementally from 2 to 5.
\begin{figure}
    \centering
    \includegraphics[width=0.4\linewidth]{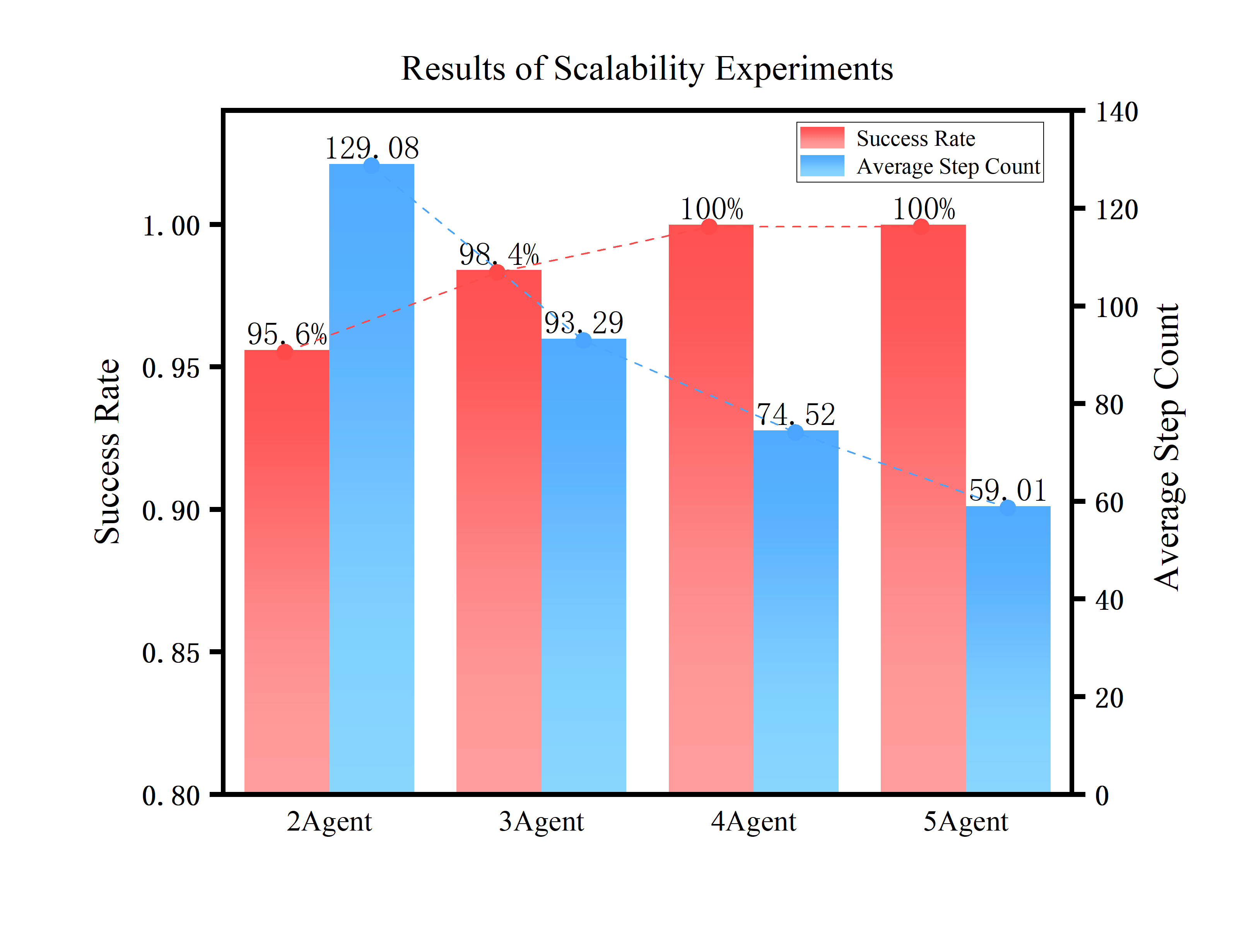}
    \caption{Effect of the number of agents on Success Rate and Average Steps in scalability experiments}
    \label{fig:7}
\end{figure}
From the experimental results, it can be observed that as the number of agents increases, the success rate steadily improves and the average number of steps decreases. Specifically, when the number of agents reaches 4, the success rate achieves 100\%, and it remains at this level when the agent count is increased to 5. Meanwhile, the average number of steps decreases significantly from 123.51 to 59.01.

These results demonstrate the excellent scalability of the PILOC framework. Even as the number of agents increases, the system maintains a high task completion rate and low search cost, highlighting its strong adaptability and potential for deployment in large-scale MASAR applications.

\section{Conclusion}
In this paper, we propose a reinforcement learning framework named PILOC, which integrates a Pheromone Inverse Guidance Mechanism and a decentralized local communication strategy to tackle the search and rescue (SAR) problem involving dynamic targets in unknown environments. This method draws inspiration from ant colony behavior by introducing pheromone-based indirect cooperation while leveraging the advantages of the CTDE architecture. Specifically, the Pheromone Inverse Guidance Mechanism enables indirect coordination in no-communication scenarios, while the local communication strategy reduces communication overhead and improves the efficiency of information sharing.

Experimental results demonstrate that PILOC achieves a significantly higher success rate, lower search cost, and stronger robustness in dynamic target environments compared to several mainstream algorithms (e.g., MASAC, IPPO, QMIX).In ablation studies, the key contributions of the Pheromone Inverse Guidance Mechanism and the local communication strategy are independently validated. Their combination synergistically enhances search efficiency and system stability. In scalability experiments, PILOC shows strong adaptability and consistent performance improvements as the number of agents increases, highlighting its potential for large-scale MASAR deployments.In summary, the proposed PILOC framework significantly enhances the cooperative SAR capability of multi-agent systems in dynamic and unknown environments, demonstrating strong adaptability and scalability.

Future work will focus on two main directions to improve the flexibility and generalization of the PILOC framework. First, we aim to expand its adaptability to different scales of task scenarios, enabling deployment across varying map sizes for broader real-world applicability. Second, we plan to incorporate more complex terrain constraints, such as areas accessible only to specific robot types. This will involve constructing heterogeneous agent systems and training specialized neural networks for different agent roles to support role-based division and personalized strategic collaboration. These enhancements will further strengthen the practicality and collaborative intelligence of the system in realistic, complex environments.

\bibliographystyle{unsrt}  
\bibliography{references}  

\begin{thebibliography}{10}

\bibitem{bahrpeyma2022review}
Fouad Bahrpeyma and Dirk Reichelt.
\newblock A review of the applications of multi-agent reinforcement learning in smart factories.
\newblock {\em Frontiers in Robotics and AI}, 9:1027340, 2022.

\bibitem{agrawal2021multi}
Akash Agrawal, Sung~Jun Won, Tushar Sharma, Mayuri Deshpande, and Christopher McComb.
\newblock A multi-agent reinforcement learning framework for intelligent manufacturing with autonomous mobile robots.
\newblock {\em Proceedings of the Design Society}, 1:161--170, 2021.

\bibitem{huang2020multi}
Yixin Huang, Shufan Wu, Zhongcheng Mu, Xiangyu Long, Sunhao Chu, and Guohong Zhao.
\newblock A multi-agent reinforcement learning method for swarm robots in space collaborative exploration.
\newblock In {\em 2020 6th international conference on control, automation and robotics (ICCAR)}, pages 139--144. IEEE, 2020.

\bibitem{yuling2023formation}
Zeng Yuling, Hao Yuqing, Yu~Ying, and Wang Qingyun.
\newblock Formation control for multi-unmanned vehicles via deep reinforcement learning.
\newblock {\em Chinese Journal of Theoretical and Applied Mechanics}, 56(2):460--471, 2023.

\bibitem{gong2023reinforcement}
Yalei Gong, Hongyun Xiong, MengMeng Li, Haibo Wang, and Xiaohong Nian.
\newblock Reinforcement learning for multi-agent formation navigation with scalability.
\newblock {\em Applied Intelligence}, 53(23):28207--28225, 2023.

\bibitem{wakilpoor2020heterogeneous}
Ceyer Wakilpoor, Patrick~J Martin, Carrie Rebhuhn, and Amanda Vu.
\newblock Heterogeneous multi-agent reinforcement learning for unknown environment mapping.
\newblock {\em arXiv preprint arXiv:2010.02663}, 2020.

\bibitem{zhang2022h2gnn}
Hao Zhang, Jiyu Cheng, Lin Zhang, Yibin Li, and Wei Zhang.
\newblock H2gnn: Hierarchical-hops graph neural networks for multi-robot exploration in unknown environments.
\newblock {\em IEEE Robotics and Automation Letters}, 7(2):3435--3442, 2022.

\bibitem{tan2022deep}
Aaron~Hao Tan, Federico~Pizarro Bejarano, Yuhan Zhu, Richard Ren, and Goldie Nejat.
\newblock Deep reinforcement learning for decentralized multi-robot exploration with macro actions.
\newblock {\em IEEE Robotics and Automation Letters}, 8(1):272--279, 2022.

\bibitem{niroui2019deep}
Farzad Niroui, Kaicheng Zhang, Zendai Kashino, and Goldie Nejat.
\newblock Deep reinforcement learning robot for search and rescue applications: Exploration in unknown cluttered environments.
\newblock {\em IEEE Robotics and Automation Letters}, 4(2):610--617, 2019.

\bibitem{de2023hierarchical}
Christopher de~Koning and Anahita Jamshidnejad.
\newblock Hierarchical integration of model predictive and fuzzy logic control for combined coverage and target-oriented search-and-rescue via robots with imperfect sensors.
\newblock {\em Journal of Intelligent \& Robotic Systems}, 107(3):40, 2023.

\bibitem{jamshidnejad2018adaptive}
Anahita Jamshidnejad and Emilio Frazzoli.
\newblock Adaptive optimal receding-horizon robot navigation via short-term policy development.
\newblock In {\em 2018 15th International Conference on Control, Automation, Robotics and Vision (ICARCV)}, pages 21--28. IEEE, 2018.

\bibitem{tang2007complete}
Fang Tang and Lynne~E Parker.
\newblock A complete methodology for generating multi-robot task solutions using asymtre-d and market-based task allocation.
\newblock In {\em Proceedings 2007 IEEE international conference on robotics and automation}, pages 3351--3358. IEEE, 2007.

\bibitem{yamauchi1998frontier}
Brian Yamauchi.
\newblock Frontier-based exploration using multiple robots.
\newblock In {\em Proceedings of the second international conference on Autonomous agents}, pages 47--53, 1998.

\bibitem{galceran2013survey}
Enric Galceran and Marc Carreras.
\newblock A survey on coverage path planning for robotics.
\newblock {\em Robotics and Autonomous systems}, 61(12):1258--1276, 2013.

\bibitem{sun2025multi}
Lijun Sun, Yijun Yang, Qiqi Duan, Yuhui Shi, Chao Lyu, Yu-Cheng Chang, Chin-Teng Lin, and Yang Shen.
\newblock Multi-agent coordination across diverse applications: A survey.
\newblock {\em arXiv preprint arXiv:2502.14743}, 2025.

\bibitem{peter2021modern}
Norvig Peter and Russell Stuart~Artificial Intelligence.
\newblock {\em A Modern Approach}.
\newblock Pearson Education, USA, 2021.

\bibitem{wang2023scrimp}
Yutong Wang, Bairan Xiang, Shinan Huang, and Guillaume Sartoretti.
\newblock Scrimp: Scalable communication for reinforcement-and imitation-learning-based multi-agent pathfinding.
\newblock In {\em 2023 IEEE/RSJ International Conference on Intelligent Robots and Systems (IROS)}, pages 9301--9308. IEEE, 2023.

\bibitem{dutta2021opportunistic}
Ayan Dutta, O~Patrick~Kreidl, and Jason~M O’Kane.
\newblock Opportunistic multi-robot environmental sampling via decentralized markov decision processes.
\newblock In {\em International Symposium Distributed Autonomous Robotic Systems}, pages 163--175. Springer, 2021.

\bibitem{farinelli2004multirobot}
Alessandro Farinelli, Luca Iocchi, and Daniele Nardi.
\newblock Multirobot systems: a classification focused on coordination.
\newblock {\em IEEE Transactions on Systems, Man, and Cybernetics, Part B (Cybernetics)}, 34(5):2015--2028, 2004.

\bibitem{gielis2022critical}
Jennifer Gielis, Ajay Shankar, and Amanda Prorok.
\newblock A critical review of communications in multi-robot systems.
\newblock {\em Current robotics reports}, 3(4):213--225, 2022.

\bibitem{ghena2019challenge}
Branden Ghena, Joshua Adkins, Longfei Shangguan, Kyle Jamieson, Philip Levis, and Prabal Dutta.
\newblock Challenge: Unlicensed lpwans are not yet the path to ubiquitous connectivity.
\newblock In {\em The 25th Annual International Conference on Mobile Computing and Networking}, pages 1--12, 2019.

\bibitem{drew2021multi}
Daniel~S Drew.
\newblock Multi-agent systems for search and rescue applications.
\newblock {\em Current Robotics Reports}, 2:189--200, 2021.

\bibitem{zhang2013coordinating}
Chongjie Zhang.~Victor Lesser.
\newblock Coordinating multi-agent reinforcement learning with limited communication.
\newblock In {\em Proceedings of the 2013 International Conference on Autonomous Agents and Multi-Agent Systems}, pages 1101--1108, 2013.

\bibitem{grupen2020low}
Niko~A Grupen, Daniel~D Lee, and Bart Selman.
\newblock Low-bandwidth communication emerges naturally in multi-agent learning systems.
\newblock {\em arXiv preprint arXiv:2011.14890}, 2020.

\bibitem{cao2024hma}
Xiao Cao, Mingyang Li, Yuting Tao, and Peng Lu.
\newblock Hma-sar: Multi-agent search and rescue for unknown located dynamic targets in completely unknown environments.
\newblock {\em IEEE Robotics and Automation Letters}, 2024.

\bibitem{chen2019self}
Fanfei Chen, Shi Bai, Tixiao Shan, and Brendan Englot.
\newblock Self-learning exploration and mapping for mobile robots via deep reinforcement learning.
\newblock In {\em Aiaa scitech 2019 forum}, page 0396, 2019.

\bibitem{lowe2017multi}
Ryan Lowe, Yi~I Wu, Aviv Tamar, Jean Harb, OpenAI Pieter~Abbeel, and Igor Mordatch.
\newblock Multi-agent actor-critic for mixed cooperative-competitive environments.
\newblock {\em Advances in neural information processing systems}, 30, 2017.

\bibitem{de2020independent}
Christian~Schroeder De~Witt, Tarun Gupta, Denys Makoviichuk, Viktor Makoviychuk, Philip~HS Torr, Mingfei Sun, and Shimon Whiteson.
\newblock Is independent learning all you need in the starcraft multi-agent challenge?
\newblock {\em arXiv preprint arXiv:2011.09533}, 2020.

\bibitem{rashid2020monotonic}
Tabish Rashid, Mikayel Samvelyan, Christian~Schroeder De~Witt, Gregory Farquhar, Jakob Foerster, and Shimon Whiteson.
\newblock Monotonic value function factorisation for deep multi-agent reinforcement learning.
\newblock {\em Journal of Machine Learning Research}, 21(178):1--51, 2020.

\end{thebibliography}

\end{document}